\documentclass[10pt,twocolumn]{article}

\usepackage{simpleConference}
\usepackage{times}
\usepackage{graphicx}
\usepackage{amsmath,amssymb}
\usepackage{url,hyperref}

\begin{document}

\title{Digital Twins for Multi-Agent Systems: When to Trust, What to Change}

\author{
\begin{tabular}{cc}
\textbf{Truc Nguyen Thi Le}$^{1,2}$
&
\textbf{Phuoc Nguyen T. H.}$^{1,2,*}$
\\[3pt]
\href{mailto:23521670@gm.uit.edu.vn}
{\small 23521670@gm.uit.edu.vn}
&
\href{mailto:phuocnth@uit.edu.vn}
{\small phuocnth@uit.edu.vn}
\end{tabular}
\\[10pt]
{\small $^1$University of Information Technology, Ho Chi Minh City, VietNam}\\[2pt]
{\small $^2$Vietnam National University Ho Chi Minh City, Ho Chi Minh City, Vietnam}\\[5pt]
{\footnotesize $^*$Corresponding author}
}

\maketitle
\thispagestyle{empty}

\begin{abstract}
Digital twins of agent networks need to flag unreliable forecasts and
help users choose interventions. Both decisions depend on what happens
over several steps: an error can spread through the network, and an edit
whose trajectory is easy to predict may still yield a poor collective
outcome. We ask when information about error propagation improves failure
detection and whether a learned twin can select effective network edits
despite imperfect forecasts. We propose Propagation-Aware Validation for
Networked Digital Twins (PAV-Twin). Its monitor combines Jacobian-based
propagation features with recent errors and ensemble disagreement to
estimate the risk that a forecast exceeds an error threshold. Its planner
simulates each candidate edit and ranks the edits by their predicted
collective outcomes. In simulations, the monitor gained
$0.13$ in area under the precision--recall curve on larger oscillator
networks with familiar topologies and $0.09$ on larger networks with
held-out topologies, relative to a graph baseline without propagation
features. Gains were small in contractive consensus. Giving the graph
baseline the same features narrowed the gap. In a separate planning study,
ensemble simulation recovered about $99\%$ of the possible improvement
over random intervention selection in oscillators and nearly all of it in
consensus. With a simplified oscillator twin, selecting by predicted
outcome also gave lower regret than selecting by trajectory fidelity.
These results show when propagation features help flag unreliable forecasts
and why the quality of intervention choices should be measured directly.
\end{abstract}

\section{Introduction}
\label{sec:Introduction}

\begin{figure}[!t]
    \centering
    \includegraphics[width=\columnwidth]{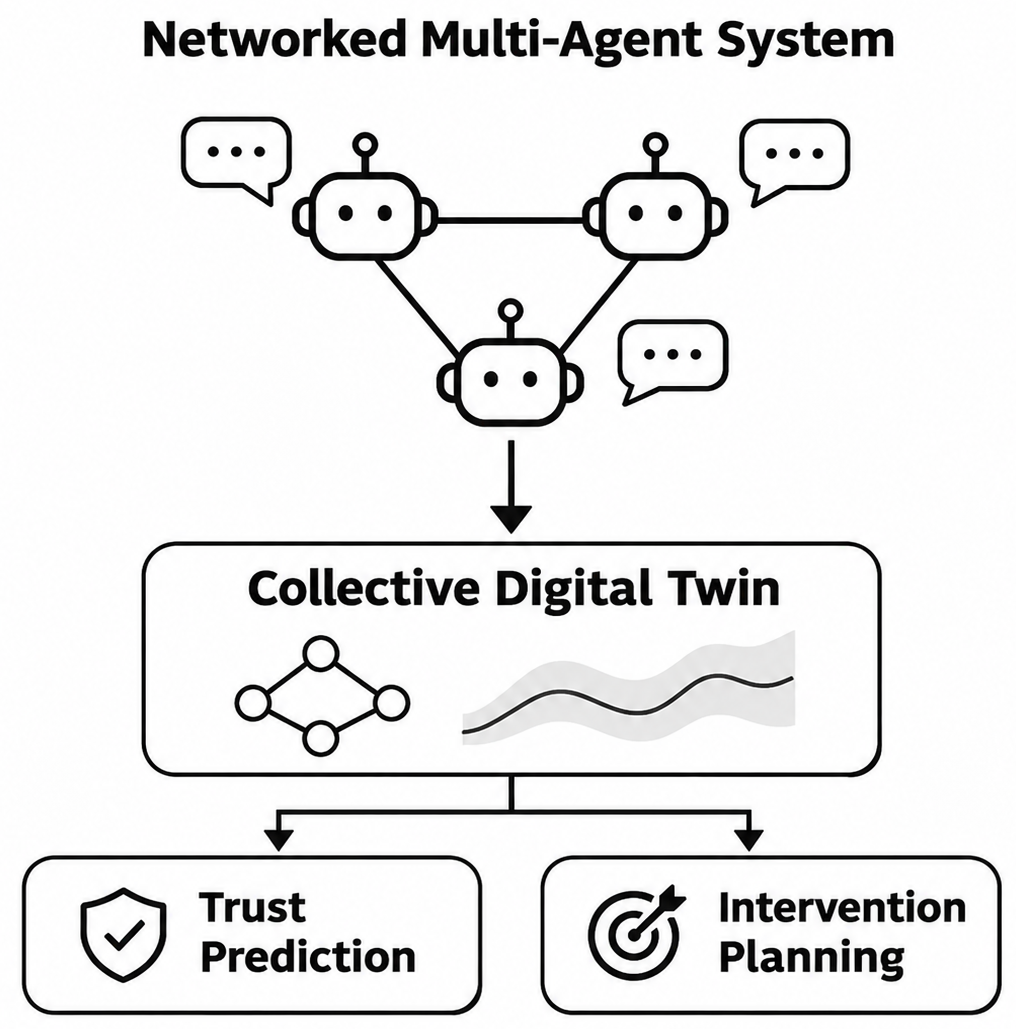}
    \caption{\textbf{Collective digital twins support forecast reliability assessment and intervention planning.}}
    \label{fig:abstract}
\end{figure}

Digital twins help people decide what to do before they alter a system
\cite{grieves2017digital,tao2019digital,fuller2020digital}. In an agent
network, that decision depends on how agents respond to one another
\cite{wooldridge1995intelligent,epstein2006generative}. Removing an edge,
for example, changes a local interaction. As neighboring agents respond,
that change can affect synchronization across the network. Prediction
errors can travel through the same interactions.
An error in one agent's predicted state can change its neighbors'
predictions and eventually distort the collective forecast. A useful twin
therefore needs to answer two questions. When is a forecast likely to
fail? Which change to the network best serves the collective goal?
Figure~\ref{fig:abstract} illustrates both tasks.

Learned world models produce multi-step forecasts by feeding each predicted
state into the next step \cite{ha2018world,janner2019trust,hafner2023mastering}.
For an agent network, a graph model gives this process a structure: nodes
represent agents and edges show who interacts with whom
\cite{battaglia2018relational}. The graph specifies the connections, and
the learned dynamics describe how each agent responds to those connections.
This gives us a way to trace how a small error can affect later states and
to simulate how the network responds when we change an edge.

For forecast reliability, recent errors and disagreement among models are
natural warning signs. They can look similar in two networks even when
the subsequent forecasts differ. One network may damp an error while the
other passes it from agent to agent. We use the twin's state Jacobians,
which measure how one predicted step responds to small changes, to estimate
how errors carry across several steps. We test whether those estimates
help detect failure as we vary coupling strength or change the network
topology. We also give a graph baseline the same features to see how much
of the observed gap remains when it has access to the same propagation
information.

Intervention planning adds a different requirement. The twin needs to
rank the candidate edits by the outcomes they produce. A twin may closely
track the trajectory after deleting one edge even though deleting a
different edge leads to better synchronization. This makes it possible
for selection by trajectory fidelity and selection by predicted outcome
to favor different edits. We compare both rules and measure the quality
of each selected edit against the best choice available in the simulator.

We address these questions with Propagation-Aware Validation for Networked
Digital Twins (PAV-Twin). We make three contributions.

\begin{enumerate}
  \item We develop a monitor that uses the twin's own dynamics to estimate
  whether a collective forecast will exceed an error threshold. The
  monitor combines estimates of propagation over several steps with
  recent errors and disagreement among twins.
  \item We test when propagation information helps. In oscillator
  networks, the gain persists on held-out topologies and peaks at an
  intermediate coupling strength in the tested sweep. Gains are small
  in contractive consensus. Giving the graph baseline the same features
  narrows the gap between monitors.
  \item We compare choosing edits by simulated outcomes with choosing by
  trajectory fidelity or supervised ranking. In the tested oscillator
  and consensus networks, ensemble planning recovers nearly all of the
  possible improvement over random selection.
\end{enumerate}

\begin{figure*}[!t]
    \centering
    \includegraphics[width=0.95\textwidth]{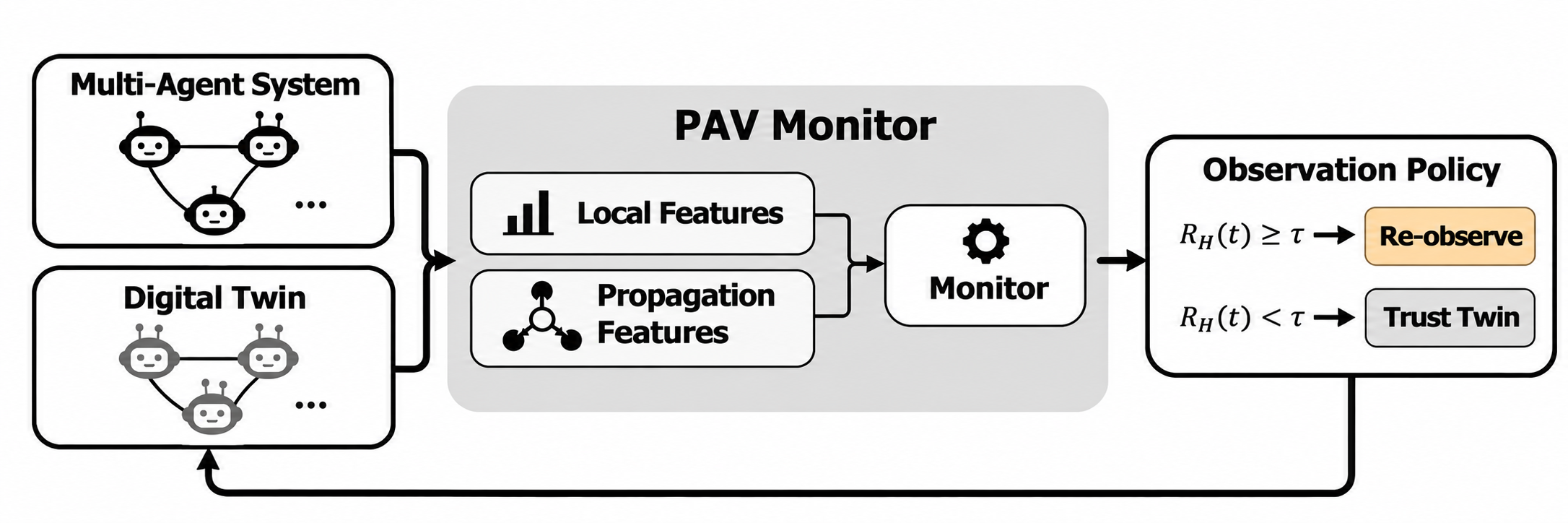}
    \caption{\textbf{PAV-Twin estimates collective forecast risk from the learned dynamics.}
    \textmd{The risk score can be used to select forecasts for re-observation.}}
    \label{fig:overview}
\end{figure*}

\section{Related Work}
\label{sec:RelatedWork}

\subsection{Digital Twins and Collective Dynamics}
\label{subsec:DigitalTwins}

Digital twins use a model of a system to predict its behavior and test
actions before they are taken \cite{grieves2017digital,tao2019digital,fuller2020digital}.
In agent-based systems, a model needs to represent how individual actions
combine to produce collective behavior
\cite{wooldridge1995intelligent,epstein2006generative}. Graph models do this
by passing information between connected agents
\cite{battaglia2018relational}. In language-model agent systems, the rules
for exchanging information also shape how agents work together. These
rules can support social simulation \cite{park2023generative}, organize
collaboration \cite{li2023camel,wu2023autogen,hong2023metagpt}, or structure
debate and consensus \cite{du2024improving,liang2023encouraging,chen2023reconcile}.
Guo et al.\ review how these systems are designed and evaluated
\cite{guo2024large}. Our study isolates the role of interactions in
simulated dynamical networks. Because the simulator specifies the agent
states and connections, we can vary the strength of those connections to
study forecast errors and test how specific edits change a collective outcome.

\subsection{Predictive Reliability and Error Propagation}
\label{subsec:MultiAgentReasoning}

Learned dynamics models make multi-step forecasts by feeding their own
predictions back into the model. Errors at one step can then affect
predictions at later steps
\cite{ha2018world,janner2019trust,hafner2023mastering}. Limiting the length
of a rollout is one way to reduce exposure to model error
\cite{janner2019trust}. For a given horizon, recent errors and
model disagreement provide evidence that a forecast may be unreliable.
Estimating where an error will spread also requires knowing how sensitive
later states are to earlier ones. This question is related to work on
how individual inputs contribute to outcomes. The Shapley value assigns
credit to players in cooperative games \cite{shapley1953value}, Aumann and
Shapley extend this idea to nonatomic games \cite{aumann1974values}, and
SHAP assigns contributions to the inputs of a prediction model
\cite{lundberg2017unified}. Our question concerns what happens over time.
We use the Jacobians of successive steps to estimate how an error at one
agent affects later states, then test whether those estimates improve
failure detection beyond recent errors and model disagreement.

\subsection{Counterfactual Intervention Planning}
\label{subsec:InterventionalPrediction}

Model-based planning compares actions by simulating what would happen
after each one \cite{janner2019trust,hafner2023mastering}. Causal models
and potential outcomes give a formal basis for comparing the outcomes of
alternative interventions \cite{pearl2009causality,peters2017elements,rubin1974estimating}.
Counterfactual policy search uses such alternatives to guide learning
\cite{buesing2018woulda}. In multi-agent reinforcement learning, the
value of one agent's action often depends on the actions of others.
Centralized critics can evaluate joint actions
\cite{lowe2017multi,foerster2018counterfactual}, and value factorization
relates team value to agent-level values \cite{sunehag2017value,rashid2018qmix}.
We study a related decision at the network level: which connections to
change. The twin simulates each candidate on the edited graph. We then
evaluate the chosen edit in the underlying simulator and measure how
close it comes to the best available outcome.

\section{Methodology}
\label{sec:Methodology}

The two tasks use the same learned dynamics. The reliability monitor uses
features extracted from the twin to estimate forecast failure risk
(Figure~\ref{fig:overview}). The intervention planner runs the twin under
candidate edits and evaluates their predicted outcomes.

\subsection{Collective Digital Twin Formulation}
\label{subsec:CollectiveTwin}

\paragraph{Simulated system.}
A networked dynamical system evolves over switching interaction graphs $G_t$,
\begin{equation}
  x_{t+1} = F_{G_t}(x_t, u_t) + w_t ,
  \label{eq:truesys}
\end{equation}
where $x_t$ stacks the agent states, $u_t$ is the control input, and $w_t$ is
process noise. We simulate nonlinear consensus, Kuramoto-like coupled
oscillators, and a coupled-map lattice (CML). Rare local disturbances and
episode-level parameter mismatch generate forecast failures that can be
difficult to anticipate from recent per-agent residuals.

\paragraph{Learned twin.}
The twin is a message-passing graph model,
\begin{equation}
  \hat{x}_{t+1} = f_\theta(x_t, A_t, u_t),
  \label{eq:twin}
\end{equation}
where $A_t$ represents $G_t$. The model learns both edge interactions and
agent-level dynamics using a multi-step rollout loss. Its analytic state
Jacobian $J_t = \partial \hat{x}_{t+1}/\partial x_t$ allows us to compute
linearized propagation features along a rollout. Computing the Jacobian
analytically avoids finite differencing the learned transition. A small
ensemble of twins provides a disagreement signal. Within each baseline
comparison, the monitors use the same underlying twin. The planning
experiments also vary twin fidelity explicitly to test its effect on
intervention choice.

\paragraph{Collective outcome.}
The planning objective $O$ assigns a score to each trajectory. For
oscillators, the score uses the Kuramoto order parameter to measure
synchronization. For consensus and CML, it uses spatial coherence
$1/(1+\mathrm{var})$, where $\mathrm{var}$ measures how much states differ
across agents. Higher values of $O$ indicate a better collective outcome.

\subsection{Forecast Reliability and Intervention Response}
\label{subsec:InterventionalModeling}

\paragraph{Reliability target.}
At forecast time $t$, the twin predicts the next $H$ states. Each window
receives a failure label if at least one prediction in that window differs
from the corresponding simulator state by more than $\epsilon$. The
monitor estimates the probability of that event,
\begin{equation}
  \begin{aligned}
    R_H(\mathcal{I}_t) &= \Pr\!\left(E_H(t) > \epsilon \mid \mathcal{I}_t\right),\\
    E_H(t) &= \max_{1\le k\le H}
        \lVert x_{t+k} - \hat{x}_{t+k}\rVert_W ,
  \end{aligned}
  \label{eq:rh}
\end{equation}
where $\lVert\cdot\rVert_W$ measures the discrepancy between predicted and
simulated states at each step. The threshold $\epsilon$ is set to a fixed
quantile of training rollout errors. Future simulator states are used to
assign labels. The monitor uses only information $\mathcal{I}_t$ available
at forecast time, including past residuals, the current state and control,
and the current topology.

\paragraph{Propagation features.}
Let $e_0$ denote the error at the start of the rollout and $e_k$ the
linearized error after step $k$. The approximation
$e_k = J_k e_{k-1} + d$ has two terms. The state Jacobian $J_k$ carries the
previous error through step $k$, while $d$ represents the local discrepancy
introduced by that step. The product $\Phi = J_H\cdots J_1$ describes how an
initial perturbation affects the state at step $H$. Its operator norm
$\lVert\Phi\rVert_2$ is the largest possible gain from the initial state to
that final step. Row summaries describe which final state components are
sensitive to initial changes. Column summaries describe which initial
components have the largest effects on the final state.

The monitor uses the linearized error forecast, the operator norm, the row
and column summaries, and per-agent sensitivities together with recent
residuals and ensemble disagreement. The final-step gain is one input to
the monitor. The failure label still depends on the largest error at any
step in the window, as in Equation~\ref{eq:rh}.

\paragraph{Intervention response.}
An intervention $I$ changes the graph $G$ to $G^I$ by adding or removing
edges, removing nodes, or scaling coupling. For each candidate, the twin
simulates a trajectory on $G^I$ and scores it using the collective
objective. We compare three choices from the same candidate set,
\begin{equation}
  \begin{aligned}
    B &= \arg\max_I O_{\mathrm{twin}}(G^I),\\
    C &= \arg\min_I \rho(G^I),\\
    A &= \arg\max_I O_{\mathrm{true}}(G^I).
  \end{aligned}
\end{equation}
The outcome-based planner $B$ chooses the edit with the highest predicted
score. Its ensemble variant is Ensemble-MPC. The fidelity-based selector
$C$ chooses the edit with the smallest trajectory tracking error $\rho$.
The simulator oracle $A$ chooses the edit with the highest realized score.
We evaluate each chosen edit in the simulator so that all choices are
compared using the same collective objective.

\subsection{Controlled Coupling Sweep}
\label{subsec:InteractionEffects}

We vary the coupling scale $\kappa \in \{0, 0.25, 0.5, 1, 2, 4\}$, holding
the other settings fixed. At $\kappa=0$, the agents evolve independently.
Contractive consensus provides a second control: agents interact, but
perturbations decay. These comparisons test whether the added value of
propagation features depends on how perturbations evolve under coupling.

\begin{figure*}[!t]
\centering
\includegraphics[width=\textwidth]{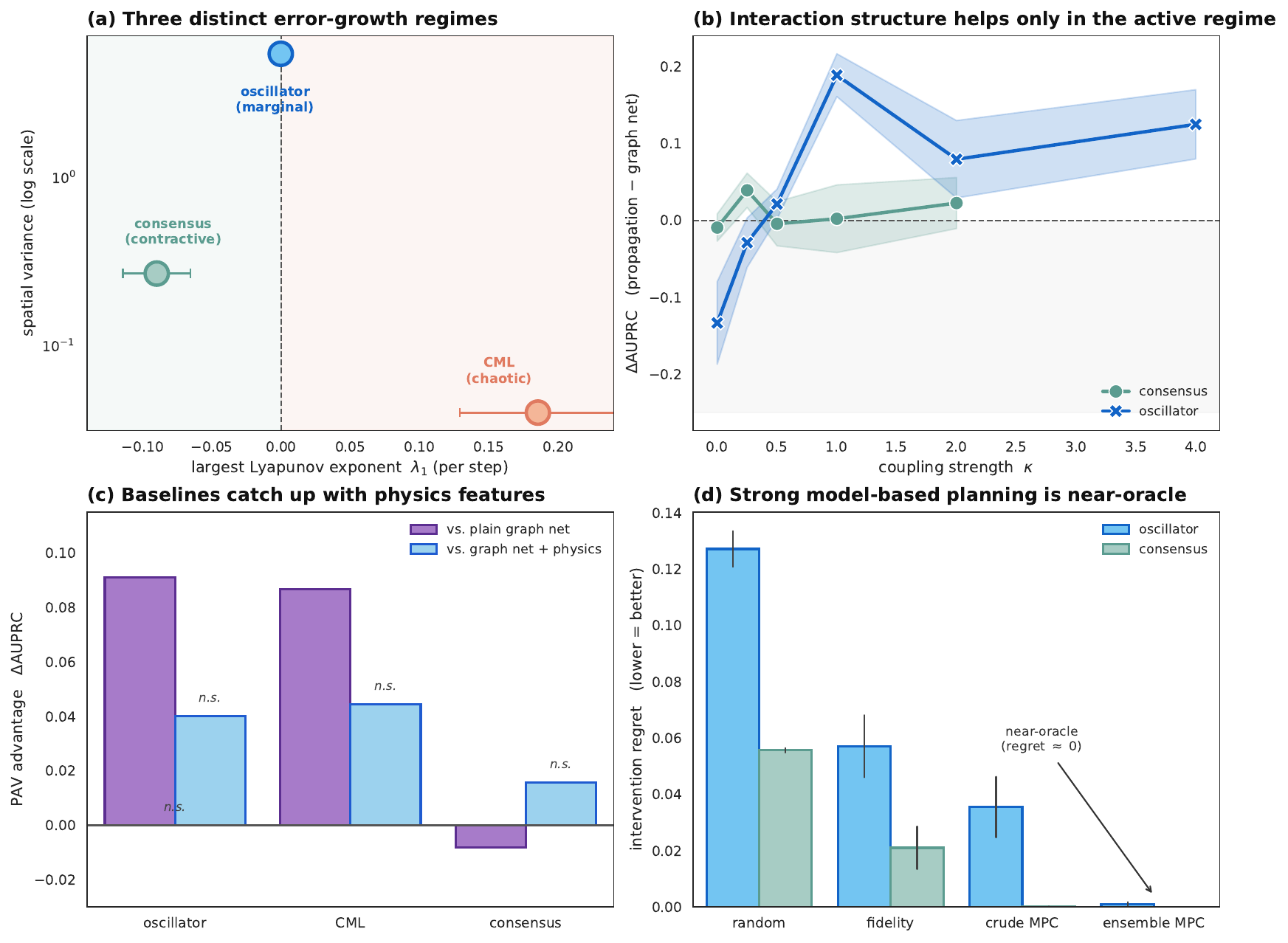}
\caption{\textbf{Forecast reliability and intervention planning across simulated dynamics.}
\textmd{\textbf{(a)} The largest Lyapunov exponent and spatial variance characterize the
three dynamics families. \textbf{(b)} The AUPRC gain over the graph baseline
varies with oscillator coupling and stays near zero in consensus.
\textbf{(c)} The gain over the graph baseline shrinks when that baseline
receives the same propagation features. \textbf{(d)} Ensemble-MPC has low
regret in oscillators and consensus. Random-selection regret measures the
steering gain available in each intervention menu.}}
\label{fig:results}
\end{figure*}

\subsection{Evaluation}
\label{subsec:EvaluationFramework}

\paragraph{Reliability prediction.}
The primary measure is area under the precision--recall curve (AUPRC) for
failure detection at horizon $H$. The protocol also considers AUROC, alarm
rates, and expected calibration error and Brier score after recalibration.
The selective policy uses the risk score to decide which fraction of
forecasts to re-observe, starting with the highest predicted risk.
The prespecified criterion for a general improvement requires a mean AUPRC
gain of at least $0.02$ over the strongest baseline in every
dynamics-by-split combination, a paired $95\%$ confidence interval with a
positive lower bound, and a win rate of at least $80\%$ across seeds.

\paragraph{Intervention planning.}
On a test split with held-out scenarios, we report regret
$O_{\mathrm{true}}(A)-O_{\mathrm{true}}(\text{selected intervention})$.
Expected regret under random selection defines the achievable
steering gain within the candidate menu. The fraction recovered is
$1-\mathrm{regret}/\mathrm{regret}_{\mathrm{random}}$ when random regret is
positive. The planning comparison includes ensemble and robust variants of
MPC and a supervised outcome ranker. Candidate interventions within a
scenario use the same noise draws, so differences in measured outcomes
reflect the graph edits under matched noise. Comparisons use paired confidence
intervals across seeds.

\section{Experiments and Results}
\label{sec:Experiments}

\begin{figure*}[!t]
\centering
\includegraphics[width=\textwidth]{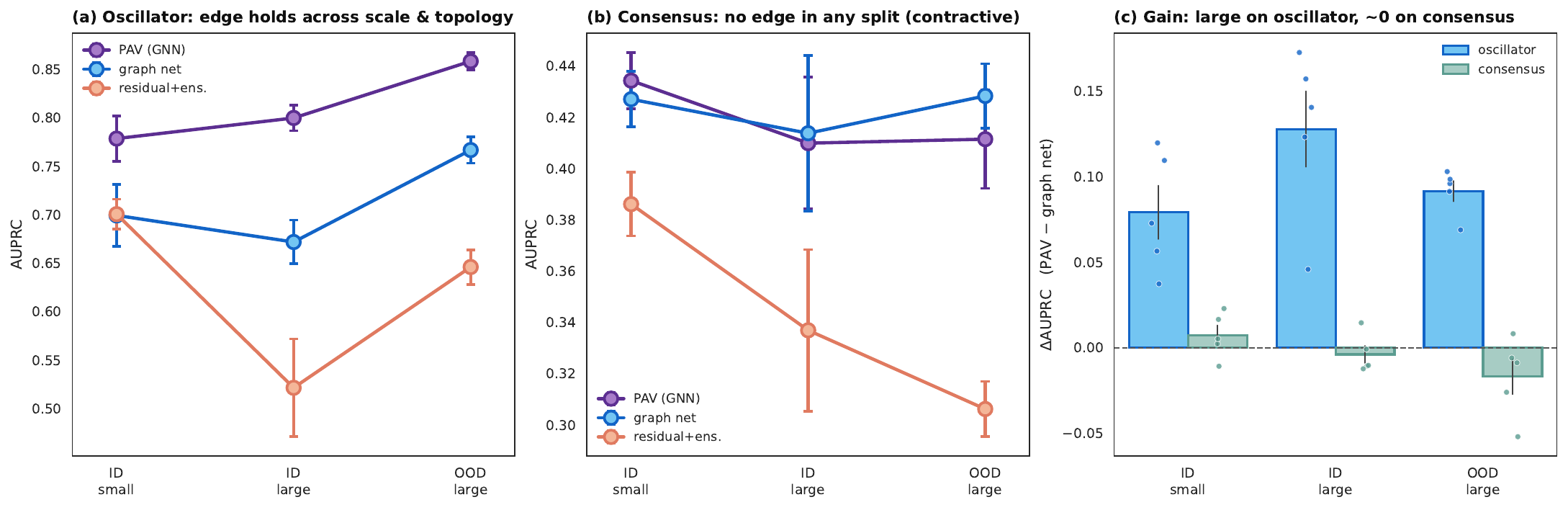}
\caption{\textbf{Failure detection across network sizes and topologies (five seeds).}
\textmd{\textbf{(a)} Oscillator AUPRC for the propagation-aware monitor, the graph
baseline, and the residual-plus-ensemble baseline. \textbf{(b)} The same
comparison in contractive consensus. \textbf{(c)} Per-split AUPRC gain over
the graph baseline, with individual runs shown as points.}}
\label{fig:transfer}
\end{figure*}

\subsection{Experimental Setup}
\label{subsec:ExperimentalSetup}

Simulations use networks of 8 to 64 agents. Training topologies include
Erd\H{o}s--R\'enyi, ring, and small-world graphs. Barab\'asi--Albert, star,
and core-periphery graphs are held out. Reliability is evaluated on networks
with familiar topologies at the training scale (\texttt{id\_small}), larger
networks with familiar topologies (\texttt{id\_large}), and larger networks
with held-out topologies (\texttt{ood\_large}). The two large-network splits
test transfer to larger networks with and without a coinciding change in
topology. We characterize the simulated dynamics independently of the
monitor. The largest Lyapunov exponent is negative in consensus, close to
zero in oscillators, and positive in the chaotic CML setting
(Table~\ref{tab:regime} and Figure~\ref{fig:results}(a)). These settings let
us study forecast errors under contraction, marginal stability, and
expansion.

\begin{table}[t]
\centering
\caption{Largest Lyapunov exponent $\lambda_1$ and spatial variance of the simulated dynamics.}
\label{tab:regime}
\small
\setlength{\tabcolsep}{4pt}
\begin{tabular*}{\columnwidth}{@{\extracolsep{\fill}}lccc@{}}
\hline
Dynamics & $\lambda_1$ (per step) & Spatial var. & Regime \\
\hline
Consensus  & $-0.090 \pm 0.024$ & $0.267$ & Contractive \\
Oscillators & $\phantom{-}0.000 \pm 0.000$ & $5.41$  & Marginally stable \\
CML        & $+0.185 \pm 0.056$ & $0.040$ & Chaotic \\
\hline
\end{tabular*}
\end{table}

\begin{figure*}[!t]
\centering
\includegraphics[width=\textwidth]{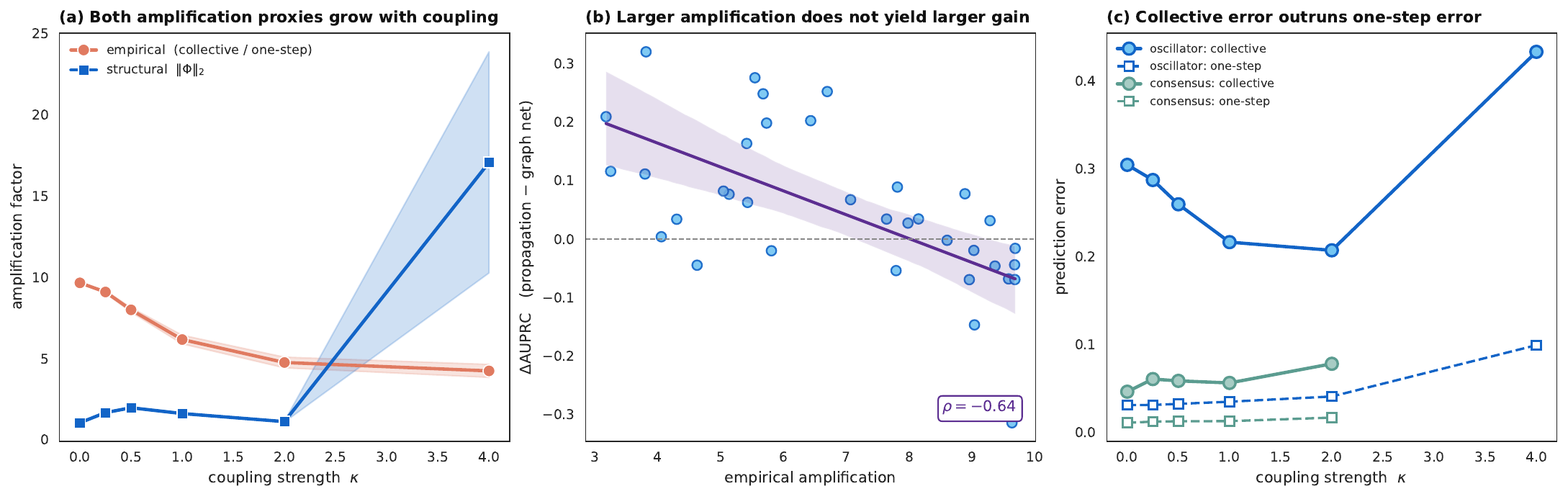}
\caption{\textbf{Error amplification and failure detection under a controlled coupling sweep.}
\textmd{\textbf{(a)} Empirical amplification is the ratio of collective rollout error
to one-step error. The operator norm $\lVert\Phi\rVert_2$ measures linearized
amplification. \textbf{(b)} AUPRC gain over the graph baseline as a function of
amplification. \textbf{(c)} One-step and collective rollout errors for
oscillators and consensus.}}
\label{fig:mech}
\end{figure*}

\subsection{Forecast Reliability Across Dynamical Regimes}
\label{subsec:InteractionEvidence}

\paragraph{Propagation improves failure detection on larger oscillator networks.}
In the five-seed transfer study, the propagation-aware monitor gained
$0.13$ AUPRC on larger oscillator networks with familiar topologies and
$0.09$ on larger networks with held-out topologies, compared with the graph
baseline without propagation features. The gain was positive on all five seeds
in both splits
(Figure~\ref{fig:transfer}(a) and Figure~\ref{fig:transfer}(c)). In the
same splits, the differences in contractive consensus were $-0.004$ and
$-0.017$ AUPRC (Figure~\ref{fig:transfer}(b) and Figure~\ref{fig:transfer}(c)).
The monitor therefore fell short of the prespecified criterion for
improvement across all dynamics and splits.

\paragraph{The oscillator gain peaks at intermediate coupling.}
In the controlled oscillator sweep, the AUPRC difference was negative when
agents were decoupled, reached about $+0.20$ near $\kappa=1$, and declined
by $\kappa=2$ (Figure~\ref{fig:results}(b)). As coupling increased, collective
rollout error outpaced one-step error in oscillators, while both errors
remained low in contractive consensus (Figure~\ref{fig:mech}(c)). The AUPRC
gain was negatively associated with the empirical error ratio and showed
little association with the operator norm (Figure~\ref{fig:mech}(b)).
Together, these observations show that the magnitude of amplification
alone gives little guidance about where propagation features improve
detection.

\paragraph{Matching propagation features narrows the gap between graph monitors.}
In a separate six-seed study, the monitor gained about $0.09$ AUPRC over the
graph baseline without propagation inputs in both oscillators and chaotic
CML. The paired $95\%$ intervals excluded zero. When a graph baseline with
the same architecture received the same features, the gap dropped to about
$0.04$ in both regimes and the intervals included zero
(Table~\ref{tab:phys} and Figure~\ref{fig:results}(c)). The reduction
supports the contribution of explicit propagation information. The extra
benefit of the proposed monitor beyond those inputs remains uncertain
relative to the matched graph baseline. The choice of comparator still
matters: the proposed monitor outperformed gradient boosting with the same
features in CML and consensus.

\begin{table}[t]
\centering
\caption{Mean AUPRC gain over each baseline across the two large-network
splits (six seeds). Asterisks indicate paired $95\%$ confidence intervals
that exclude zero.}
\label{tab:phys}
\small
\setlength{\tabcolsep}{4pt}
\begin{tabular*}{\columnwidth}{@{\extracolsep{\fill}}lccc@{}}
\hline
Dynamics & Graph & \shortstack{Graph +\\ propagation} &
\shortstack{Boosting +\\ propagation} \\
\hline
Oscillators & $+0.091^{\ast}$ & $+0.040$ & $+0.022$ \\
CML         & $+0.087^{\ast}$ & $+0.044$ & $+0.186^{\ast}$ \\
Consensus   & $-0.008$       & $+0.016$ & $+0.072^{\ast}$ \\
\hline
\end{tabular*}
\end{table}

\begin{figure*}[!t]
\centering
\includegraphics[width=\textwidth]{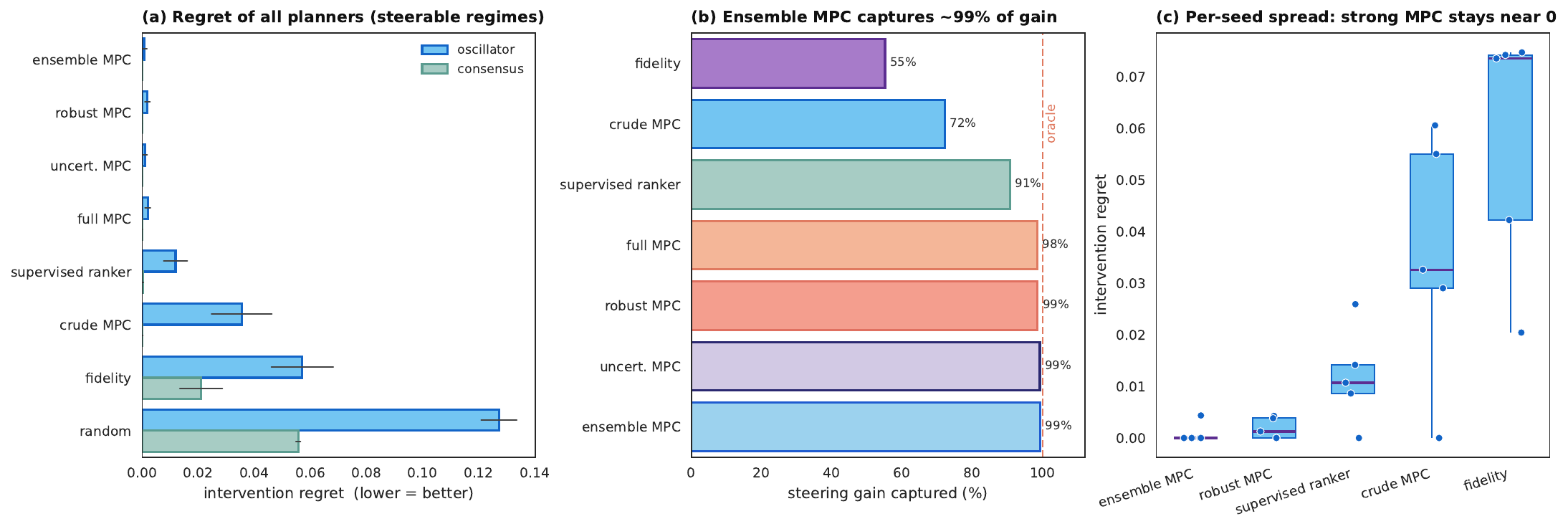}
\caption{\textbf{Intervention planning in simulated oscillators and consensus (five seeds).}
\textmd{\textbf{(a)} Regret by planning method. \textbf{(b)} Fraction of the
achievable steering gain recovered in oscillators. \textbf{(c)} Per-seed
regret for selected methods.}}
\label{fig:planning}
\end{figure*}

\subsection{Trajectory Fidelity and Intervention Quality}
\label{subsec:ResponsePrediction}

In oscillators, selecting by predicted synchronization yielded lower regret
than selecting by trajectory fidelity with both the full and simplified
twins (Table~\ref{tab:plan}). With the simplified twin, outcome-based regret
was about $89\%$ lower than fidelity-based regret, and fidelity-based
selection performed worse than random. The rankings show where the selectors
diverged. With the simplified twin, the outcome-based ranking had a positive
rank correlation of $0.70$ with true outcomes, while the fidelity-based
ranking had a correlation of $-0.76$.

The difference between selectors was smaller in consensus. Outcome-based
selection reached oracle-level regret at the reported precision with both
twins, and fidelity-based selection did so with the simplified twin
(Table~\ref{tab:plan}). The oscillator results therefore establish a failure
mode for fidelity-based selection, while the consensus results bound the
scope of that finding.

\begin{table}[t]
\centering
\caption{Intervention regret for outcome-based selection ($B$), fidelity-based
selection ($C$), and random selection. Lower is better.}
\label{tab:plan}
\small
\setlength{\tabcolsep}{4pt}
\begin{tabular*}{\columnwidth}{@{\extracolsep{\fill}}llccc@{}}
\hline
Dynamics & Twin & Outcome & Fidelity & Random \\
\hline
Oscillators & Full       & $0.001$ & $0.056$ & $0.090$ \\
Oscillators & Simplified & $0.019$ & $0.173$ & $0.090$ \\
Consensus   & Full       & $0.000$ & $0.023$ & $0.043$ \\
Consensus   & Simplified & $0.000$ & $0.000$ & $0.043$ \\
\hline
\end{tabular*}
\end{table}

\subsection{Intervention Planning Baselines}
\label{subsec:ControlledDynamics}

\paragraph{Ensemble planning approaches the oracle in oscillators and consensus.}
In a separate five-seed study, Ensemble-MPC recovered approximately $99\%$
of the available steering gain in oscillators and nearly all of it in
consensus (Table~\ref{tab:stage1}, Figure~\ref{fig:planning}(b), and
Figure~\ref{fig:results}(d)). In oscillators, its mean regret was around
$97\%$ lower than that of MPC with a single simplified twin. Ensemble and
robust MPC also showed little variation across seeds
(Figure~\ref{fig:planning}(a) and Figure~\ref{fig:planning}(c)).

The supervised outcome ranker had higher mean oscillator regret than
Ensemble-MPC. The paired difference was $+0.011$ with a $95\%$ confidence
interval of $[-0.001,\,0.023]$, which includes zero
(Table~\ref{tab:stage1}). In chaotic and periodic CML, random
regret was already small. The candidate edits spanned about $1\%$ of the
outcome range and at most $5\%$, leaving little scope for distinguishing
planners by absolute regret.

\begin{table}[t]
\centering
\caption{Mean planning regret in the five-seed study. The random regret in
parentheses measures the steering gain available within each candidate menu.}
\label{tab:stage1}
\small
\setlength{\tabcolsep}{3pt}
\begin{tabular*}{\columnwidth}{@{\extracolsep{\fill}}lcccc@{}}
\hline
\shortstack[l]{Regime\\ (random regret)} &
\shortstack{Ensemble\\ MPC} & \shortstack{Simplified\\ MPC} &
Ranker & Fidelity \\
\hline
Oscillators ($0.127$)  & $0.0009$ & $0.0355$ & $0.0119$ & $0.057$ \\
Consensus ($0.056$)    & $0.0000$ & $0.0001$ & $0.0002$ & $0.021$ \\
Chaotic CML ($0.007$)  & $0.006$  & $0.006$  & $0.005$  & $0.003$ \\
Periodic CML ($0.002$) & $0.001$  & $0.001$  & $0.001$  & $0.001$ \\
\hline
\end{tabular*}
\end{table}

\section{Conclusion}
\label{sec:Conclusion}

Propagation features improved collective failure detection in the simulated
oscillator and chaotic CML settings relative to a graph baseline without
those features. The oscillator sweep and the contractive consensus results
show that the benefit depends on the dynamics. For steering, ensemble
simulation yielded near-oracle choices in the tested oscillator and consensus
settings. In oscillators, evaluating interventions by their predicted
collective outcomes also gave consistently lower regret than selecting the
trajectories the twin could track most accurately.

\section{Limitations and Future Work}
\label{subsec:Limitations}

The transfer results measure how well the monitors rank forecast risk.
Choosing when to re-observe a system also depends on the calibration of
its predicted probabilities. Whether that calibration holds when coupling
shifts beyond the training range is a separate question.

The reported win rates summarize a small number of seeds. Larger replications
could use score intervals for the probability of a seed-level win
\cite{wilson1927probable} and paired bootstrap intervals to assess the
sensitivity of mean performance differences to the sampled seeds
\cite{efron1993introduction}.

The feature-matched comparison leaves open whether a learned representation
of propagation could improve on the Jacobian summaries used here. The
planning results are likewise bounded by the simulated networks and the
candidate graph edits. Broader intervention menus would give a stronger test
of planning in chaotic dynamics, where the menus used here offered little
steering gain. Higher-dimensional systems would also let us test when the
twin's outcome ranking becomes unreliable and whether an intervention ranker
can help in those settings.

\bibliographystyle{unsrt}
\bibliography{references}

@incollection{grieves2017digital,
  author = {Grieves, Michael and Vickers, John},
  title = {Digital Twin: Mitigating Unpredictable, Undesirable Emergent Behavior in Complex Systems},
  booktitle = {Transdisciplinary Perspectives on Complex Systems},
  publisher = {Springer International Publishing},
  year = {2017},
  doi = {10.1007/978-3-319-38756-7_4},
  pages = {85--113},
}

@article{tao2019digital,
  author = {Tao, Fei and Zhang, He and Liu, Ang and Nee, A. Y. C.},
  title = {Digital Twin in Industry: State-of-the-Art},
  journal = {IEEE Transactions on Industrial Informatics},
  volume = {15},
  number = {4},
  pages = {2405--2415},
  year = {2019},
  doi = {10.1109/TII.2018.2873186},
  publisher = {Institute of Electrical and Electronics Engineers (IEEE)},
}

@article{fuller2020digital,
  author = {Fuller, Aidan and Fan, Zhong and Day, Charles and Barlow, Chris},
  title = {Digital Twin: Enabling Technologies, Challenges and Open Research},
  journal = {IEEE Access},
  volume = {8},
  pages = {108952--108971},
  year = {2020},
  doi = {10.1109/ACCESS.2020.2998358},
  publisher = {Institute of Electrical and Electronics Engineers (IEEE)},
}

@article{wooldridge1995intelligent,
  author = {Wooldridge, Michael and Jennings, Nicholas R.},
  title = {Intelligent Agents: Theory and Practice},
  journal = {The Knowledge Engineering Review},
  volume = {10},
  number = {2},
  pages = {115--152},
  year = {1995},
  doi = {10.1017/S0269888900008122},
  publisher = {Cambridge University Press},
}

@book{epstein2006generative,
  author = {Epstein, Joshua M.},
  title = {Generative Social Science: Studies in Agent-Based Computational Modeling},
  publisher = {Princeton University Press},
  year = {2006},
  url = {https://press.princeton.edu/books/ebook/9781400842872/generative-social-science-0},
}

@misc{ha2018world,
  author = {Ha, David and Schmidhuber, J{\"u}rgen},
  title = {World Models},
  year = {2018},
  eprint = {1803.10122},
  archiveprefix = {arXiv},
  doi = {10.5281/zenodo.1207631},
  howpublished = {\url{https://worldmodels.github.io/}},
}

@inproceedings{janner2019trust,
  author = {Janner, Michael and Fu, Justin and Zhang, Marvin and Levine, Sergey},
  title = {When to Trust Your Model: Model-Based Policy Optimization},
  booktitle = {Neural Information Processing Systems},
  year = {2019},
  eprint = {1906.08253},
  archiveprefix = {arXiv},
  doi = {10.48550/arXiv.1906.08253},
}

@article{hafner2023mastering,
  author = {Hafner, Danijar and Pasukonis, Jurgis and Ba, Jimmy and Lillicrap, Timothy},
  title = {Mastering Diverse Control Tasks through World Models},
  year = {2025},
  eprint = {2301.04104},
  archiveprefix = {arXiv},
  url = {https://www.nature.com/articles/s41586-025-08744-2},
  journal = {Nature},
  doi = {10.1038/s41586-025-08744-2},
  volume = {640},
  number = {8059},
  pages = {647--653},
}

@misc{battaglia2018relational,
  author = {Battaglia, P. and Hamrick, Jessica B. and Bapst, V. and Sanchez-Gonzalez, Alvaro and Zambaldi, V. and Malinowski, Mateusz and Tacchetti, Andrea and Raposo, D. and Santoro, A. and Faulkner, Ryan and others},
  title = {Relational Inductive Biases, Deep Learning, and Graph Networks},
  year = {2018},
  eprint = {1806.01261},
  archiveprefix = {arXiv},
  doi = {10.48550/arXiv.1806.01261},
  howpublished = {arXiv:1806.01261},
  url = {https://arxiv.org/abs/1806.01261},
}

@inproceedings{park2023generative,
  author = {Park, Joon Sung and O'Brien, Joseph C. and Cai, Carrie J. and Morris, Meredith Ringel and Liang, Percy and Bernstein, Michael S.},
  title = {Generative Agents: Interactive Simulacra of Human Behavior},
  booktitle = {ACM Symposium on User Interface Software and Technology},
  year = {2023},
  doi = {10.1145/3586183.3606763},
  pages = {1--22},
  publisher = {ACM},
}

@inproceedings{li2023camel,
  author = {Li, Guohao and Hammoud, Hasan Abed Al Kader and Itani, Hani and Khizbullin, Dmitrii and Ghanem, Bernard},
  title = {{CAMEL}: Communicative Agents for {``Mind''} Exploration of Large Language Model Society},
  year = {2023},
  eprint = {2303.17760},
  archiveprefix = {arXiv},
  url = {https://proceedings.neurips.cc/paper_files/paper/2023/hash/a3621ee907def47c1b952ade25c67698-Abstract-Conference.html},
  pages = {51991--52008},
  doi = {10.52202/075280-2264},
  publisher = {Neural Information Processing Systems Foundation, Inc. (NeurIPS)},
  booktitle = {Advances in Neural Information Processing Systems},
  volume = {36},
}

@inproceedings{wu2023autogen,
  author = {Wu, Qingyun and Bansal, Gagan and Zhang, Jieyu and Wu, Yiran and Li, Beibin and Zhu, Erkang and Jiang, Li and Zhang, Xiaoyun and Zhang, Shaokun and Liu, Jiale and Awadallah, Ahmed Hassan and White, Ryen W. and Burger, Doug and Wang, Chi},
  title = {{AutoGen}: Enabling Next-Gen {LLM} Applications via Multi-Agent Conversations},
  year = {2024},
  eprint = {2308.08155},
  archiveprefix = {arXiv},
  url = {https://openreview.net/forum?id=BAakY1hNKS},
  booktitle = {Proceedings of the First Conference on Language Modeling},
}

@inproceedings{hong2023metagpt,
  author = {Hong, Sirui and Zhuge, Mingchen and Chen, Jonathan and Zheng, Xiawu and Cheng, Yuheng and Zhang, Ceyao and Wang, Jinlin and Wang, Zili and Yau, Steven Ka Shing and Lin, Zijuan and Zhou, Liyang and Ran, Chenyu and Xiao, Lingfeng and Wu, Chenglin and Schmidhuber, J{\"u}rgen},
  title = {{MetaGPT}: Meta Programming for a Multi-Agent Collaborative Framework},
  year = {2024},
  eprint = {2308.00352},
  archiveprefix = {arXiv},
  url = {https://proceedings.iclr.cc/paper_files/paper/2024/file/6507b115562bb0a305f1958ccc87355a-Paper-Conference.pdf},
  booktitle = {International Conference on Learning Representations},
}

@inproceedings{du2024improving,
  author = {Du, Yilun and Li, Shuang and Torralba, Antonio and Tenenbaum, Joshua B. and Mordatch, Igor},
  title = {Improving Factuality and Reasoning in Language Models through Multiagent Debate},
  booktitle = {Proceedings of the 41st International Conference on Machine Learning},
  series = {Proceedings of Machine Learning Research},
  volume = {235},
  year = {2024},
  url = {https://proceedings.mlr.press/v235/du24e.html},
  doi = {10.48550/arXiv.2305.14325},
  pages = {11733--11763},
}

@inproceedings{liang2023encouraging,
  author = {Liang, Tian and He, Zhiwei and Jiao, Wenxiang and Wang, Xing and Wang, Yan and Wang, Rui and Yang, Yujiu and Shi, Shu-Ming and Tu, Zhaopeng},
  title = {Encouraging Divergent Thinking in Large Language Models through Multi-Agent Debate},
  year = {2024},
  eprint = {2305.19118},
  archiveprefix = {arXiv},
  url = {https://aclanthology.org/2024.emnlp-main.992/},
  pages = {17889--17904},
  doi = {10.18653/v1/2024.emnlp-main.992},
  publisher = {Association for Computational Linguistics},
  booktitle = {Proceedings of the 2024 Conference on Empirical Methods in Natural Language Processing},
}

@inproceedings{chen2023reconcile,
  author = {Chen, Justin and Saha, Swarnadeep and Bansal, Mohit},
  title = {{ReConcile}: Round-Table Conference Improves Reasoning via Consensus among Diverse {LLMs}},
  year = {2024},
  eprint = {2309.13007},
  archiveprefix = {arXiv},
  url = {https://aclanthology.org/2024.acl-long.381/},
  pages = {7066--7085},
  doi = {10.18653/v1/2024.acl-long.381},
  publisher = {Association for Computational Linguistics},
  booktitle = {Proceedings of the 62nd Annual Meeting of the Association for Computational Linguistics (Volume 1: Long Papers)},
}

@inproceedings{guo2024large,
  author = {Guo, Taicheng and Chen, Xiuying and Wang, Yaqi and Chang, Ruidi and Pei, Shichao and Chawla, Nitesh V. and Wiest, Olaf and Zhang, Xiangliang},
  title = {Large Language Model Based Multi-agents: A Survey of Progress and Challenges},
  booktitle = {Proceedings of the Thirty-Third International Joint Conference on Artificial Intelligence},
  year = {2024},
  doi = {10.24963/ijcai.2024/890},
  pages = {8048--8057},
  url = {https://www.ijcai.org/proceedings/2024/890},
}

@incollection{shapley1953value,
  author = {Shapley, Lloyd S.},
  title = {A Value for n-Person Games},
  booktitle = {Contributions to the Theory of Games II},
  publisher = {Princeton University Press},
  year = {1953},
  pages = {307--317},
}

@book{aumann1974values,
  author = {Aumann, Robert J. and Shapley, Lloyd S.},
  title = {Values of Non-Atomic Games},
  publisher = {Princeton University Press},
  year = {1974},
}

@inproceedings{lundberg2017unified,
  author = {Lundberg, Scott M. and Lee, Su-In},
  title = {A Unified Approach to Interpreting Model Predictions},
  booktitle = {Advances in Neural Information Processing Systems},
  year = {2017},
  volume = {30},
  url = {https://proceedings.neurips.cc/paper_files/paper/2017/hash/8a20a8621978632d76c43dfd28b67767-Abstract.html},
}

@book{pearl2009causality,
  author = {Pearl, Judea},
  title = {Causality: Models, Reasoning, and Inference},
  edition = {Second},
  publisher = {Cambridge University Press},
  year = {2009},
}

@book{peters2017elements,
  author = {Peters, J. and Janzing, D. and Schölkopf, B.},
  title = {Elements of Causal Inference: Foundations and Learning Algorithms},
  publisher = {MIT Press},
  year = {2017},
}

@article{rubin1974estimating,
  author = {Rubin, Donald B.},
  title = {Estimating Causal Effects of Treatments in Randomized and Nonrandomized Studies},
  journal = {Journal of Educational Psychology},
  volume = {66},
  number = {5},
  pages = {688--701},
  year = {1974},
  doi = {10.1037/h0037350},
  publisher = {American Psychological Association (APA)},
}

@misc{buesing2018woulda,
  author = {Buesing, L. and Weber, T. and Zwols, Yori and Racanière, S. and Guez, A. and Lespiau, Jean-Baptiste and Heess, N.},
  title = {Woulda, Coulda, Shoulda: Counterfactually-Guided Policy Search},
  year = {2018},
  eprint = {1811.06272},
  archiveprefix = {arXiv},
  doi = {10.48550/arXiv.1811.06272},
  howpublished = {arXiv:1811.06272},
  url = {https://arxiv.org/abs/1811.06272},
}

@inproceedings{lowe2017multi,
  author = {Lowe, Ryan and Wu, Yi and Tamar, Aviv and Harb, Jean and Abbeel, Pieter and Mordatch, Igor},
  title = {Multi-Agent Actor-Critic for Mixed Cooperative-Competitive Environments},
  booktitle = {Neural Information Processing Systems},
  year = {2017},
  doi = {10.48550/arXiv.1706.02275},
}

@inproceedings{foerster2018counterfactual,
  author = {Foerster, Jakob and Farquhar, Gregory and Afouras, Triantafyllos and Nardelli, Nantas and Whiteson, Shimon},
  title = {Counterfactual Multi-Agent Policy Gradients},
  booktitle = {Proceedings of the AAAI Conference on Artificial Intelligence},
  volume = {32},
  number = {1},
  year = {2018},
  doi = {10.1609/aaai.v32i1.11794},
  publisher = {Association for the Advancement of Artificial Intelligence (AAAI)},
}

@inproceedings{sunehag2017value,
  author = {Sunehag, Peter and Lever, Guy and Gruslys, Audrunas and Czarnecki, Wojciech Marian and Zambaldi, Vinicius and Jaderberg, Max and Lanctot, Marc and Sonnerat, Nicolas and Leibo, Joel Z. and Tuyls, Karl and Graepel, Thore},
  title = {Value-Decomposition Networks for Cooperative Multi-Agent Learning Based on Team Reward},
  year = {2018},
  eprint = {1706.05296},
  archiveprefix = {arXiv},
  pages = {2085--2087},
  doi = {10.65109/jsrc7365},
  publisher = {International Foundation for Autonomous Agents and Multiagent Systems},
  booktitle = {Proceedings of the 17th International Conference on Autonomous Agents and Multiagent Systems},
  url = {https://www.ifaamas.org/Proceedings/aamas2018/pdfs/p2085.pdf},
}

@inproceedings{rashid2018qmix,
  author = {Rashid, Tabish and Samvelyan, Mikayel and Schroeder, Christian and Farquhar, Gregory and Foerster, Jakob and Whiteson, Shimon},
  title = {{QMIX}: Monotonic Value Function Factorisation for Deep Multi-Agent Reinforcement Learning},
  booktitle = {Proceedings of the 35th International Conference on Machine Learning},
  year = {2018},
  url = {https://proceedings.mlr.press/v80/rashid18a.html},
  series = {Proceedings of Machine Learning Research},
  volume = {80},
  pages = {4295--4304},
}

@article{wilson1927probable,
  author = {Wilson, Edwin B.},
  title = {Probable Inference, the Law of Succession, and Statistical Inference},
  journal = {Journal of the American Statistical Association},
  volume = {22},
  number = {158},
  pages = {209--212},
  year = {1927},
  doi = {10.1080/01621459.1927.10502953},
  publisher = {Informa UK Limited},
}

@book{efron1993introduction,
  author = {Efron, Bradley and Tibshirani, Robert J.},
  title = {An Introduction to the Bootstrap},
  publisher = {Chapman and Hall/CRC},
  year = {1993},
}

\end{document}